\documentclass[12pt,journal]{IEEEtran}
\usepackage{amssymb,amsmath}
\usepackage{amssymb,amsmath}
\usepackage{setspace}
\usepackage{graphicx}
\usepackage{epsfig,cite,latexsym,subfigure}
\usepackage{theorem}
\usepackage{times}
\usepackage{epsf,cite,subfigure}
\usepackage{amsmath,amssymb}

\usepackage{floatflt}
\usepackage{url}
\usepackage{times}
\usepackage[usenames]{color}
\usepackage{dsfont}
\usepackage{mathptmx}
\usepackage[final]{pdfpages}
\newcommand{\set}[1]{\left[ #1\right]}

\newcommand     {\curls}[1]{\left\{ #1\right\} }
\newcommand     {\paren}[1]{\left(#1\right)}
\newcommand{\C}{\mathbb{C}} 

\newcommand{\eqnlabel}[1]{\label{eqn:#1}}
\newcommand{\eqnref}[1]{(\ref{eqn:#1})}

\title{{\Large  Reservoir-Based Distributed Machine Learning\\ for Edge Operation}}\vspace{-2mm}
\author{
\IEEEauthorblockN{\normalsize Silvija Kokalj-Filipovic, Paul Toliver, William Johnson, Rob Miller} \\ \vspace{-2mm} \vspace{\baselineskip} 
\IEEEauthorblockA{\small Perspecta Labs Inc  \vspace{\baselineskip} 
\small\em \{skfilipovic, ptoliver, wjohnson, rmiller\}@perspectalabs.com}}
\begin{document}
\maketitle
\begin{abstract}
We introduce a novel design for in-situ
training of machine learning algorithms built into smart
sensors, and illustrate distributed training scenarios using radio
frequency (RF) spectrum sensors. Current RF sensors at the
Edge lack the computational resources to support practical, in-situ
training for intelligent signal classification. We propose a
solution using Deep-delay Loop Reservoir Computing (DLR),
a processing architecture that supports machine learning algorithms
on resource-constrained edge-devices 
by leveraging delay-loop reservoir computing in combination
with innovative hardware. 
DLR delivers reductions in form factor,
hardware complexity and latency, compared to the State-of-the-
Art (SoA) neural nets. We demonstrate 
DLR for two applications: RF Specific Emitter Identification
(SEI) and wireless protocol recognition. DLR enables mobile
edge platforms to authenticate and then track emitters with fast SEI retraining. Once delay loops separate the
data classes, traditionally complex, power-hungry classification
models are no longer needed for the learning process. Yet,
even with simple classifiers such as Ridge Regression (RR),
the complexity grows at least quadratically with the input size. DLR with a RR classifier
exceeds the SoA accuracy, while further reducing
power consumption by leveraging the architecture of parallel
(split) loops. To authenticate mobile devices across large regions,
DLR can be trained in a distributed fashion with very little
additional processing and a small communication cost, all while
maintaining accuracy. 
We illustrate how to merge locally trained DLR classifiers in use cases of interest.
\end{abstract}
\section{Introduction}
\begin{figure}[h]
\vspace{-1mm}
\centering
\includegraphics[width=0.49\textwidth]{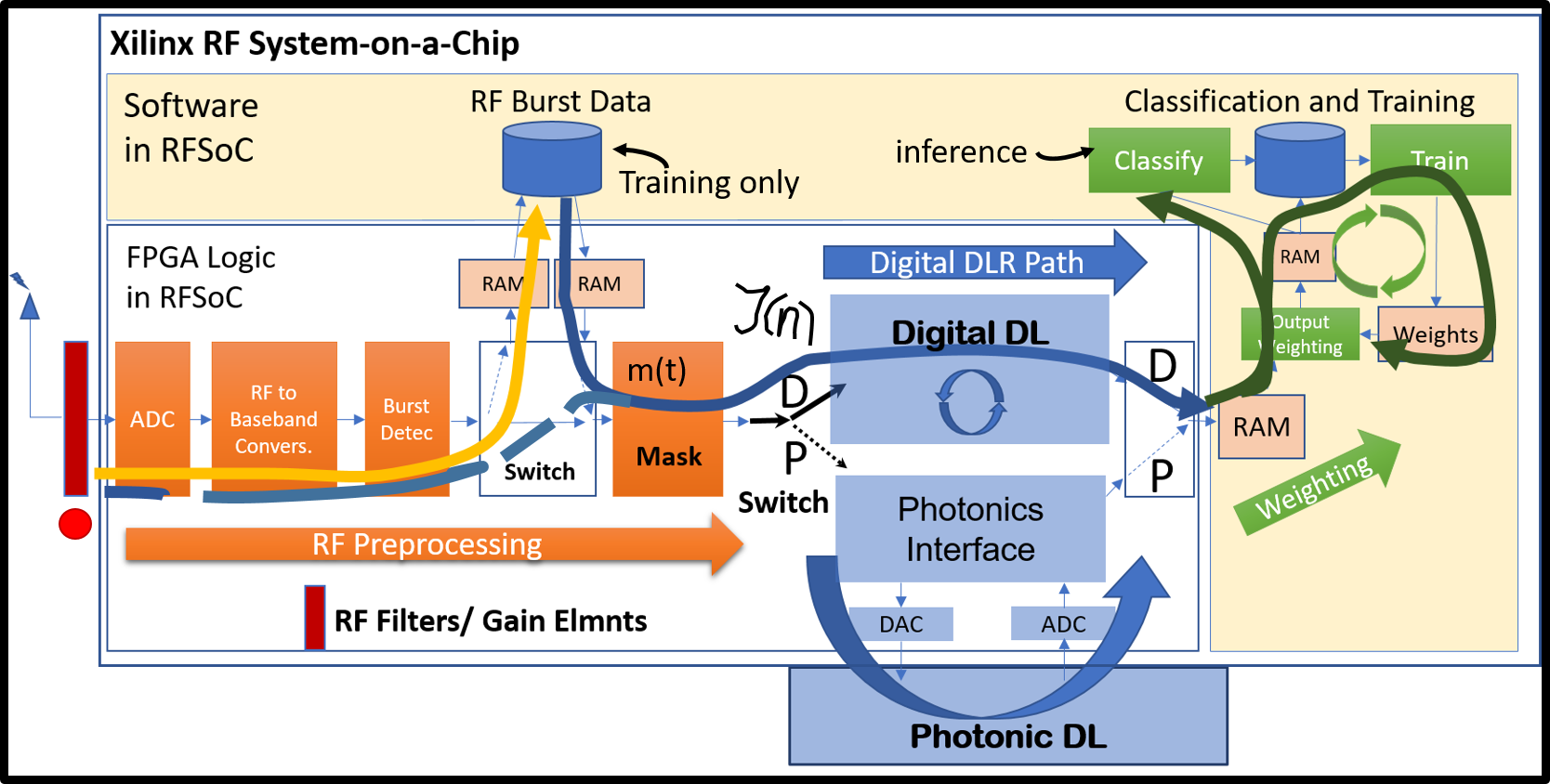}
\vspace{-3mm}
\caption{The DLR demo platform performing the in-situ SEI/WiPRec training and inference with two selectable delay loop implementations (D, for digital, and P for photonics). The yellow path is application specific (RF data ingest and burst extraction), while the full-line blue path is generic across applications: burst of samples $~\longmapsto ~$ delay loops $~\longmapsto ~$ state vector $~\longmapsto ~$ RR training or classification. The dash-lined blue path shows how the data ingest connects to the loop during the inference phase.}\vspace{-1mm}
\label{fig:platform}
\end{figure}

Edge computing is a paradigm where computing is distributed across a large number of small devices instead of being centralized in the Cloud. Frequently,  edge computing occurs near the physical location where data is being collected and analyzed. However, not all computational algorithms are readily adaptable to the Edge even if the data is produced there.
State-of-the-Art (SoA) machine learning systems that are trained on sensor signals lack the computational resources to support in-situ training and adaptable inference for situational awareness. Such in-situ solutions are needed as it is not always practical to leverage backhaul resources due to security, bandwidth, and mission latency requirements. We propose a solution through Deep-delay Loop Reservoir Computing (DLR), our novel AI processing architecture that supports retrainable machine learning (ML) solutions on resource constrained edge-devices (e.g. mobile phones) by leveraging delay-loop reservoir computing (RC). Our prototype DLR platform \cite{RFSoC} supports two loop implementations using a configurable switch (see Fig.~\ref{fig:platform}): a digital (FPGA) loop, and the analogue one, based on the innovative photonic hardware that exploits the inherent speed and multi-dimensional (spatial, temporal and wavelength-based) processing diversity of signals in the optical domain \cite{AdvancesinphotonicRC}. Reservoir computing is a bio-inspired approach especially suited for processing time-dependent information in a computationally efficient way \cite{Lukosevicius2009ReservoirCA} in order to facilitate the learning from such information. The RC in ML solutions conditions the input features towards linear separability of different classes, upon which a simple linear ML algorithm can be trained to achieve high accuracy of classification. Such a simple ML model can be trained efficiently, and that is one of the reasons why DLR delivers significant reductions in the form factor and hardware complexity/  power consumption for training at the Edge, providing real-time learning latency that is several orders of magnitude lower than with the SoA classifiers. We demonstrate the advantages of DLR on the applications of RF Specific Emitter Identification (SEI) and wireless protocol recognition. SEI  aims to extract rich nonlinear characteristics of internal components within a transmitter to distinguish one transmitter from another, even within the same manufacturer and protocol class. For an illustration of the DLR system used  for SEI please see Fig.~\ref{fig:DLRsys} or consult our prior work \cite{GomacTech} that reports intermediate results and describes the preliminary architecture of the photonic loop, which is not covered here. We have since improved the performance of the DLR as well as the models of the SoA classifiers performing the SEI on the signal bursts emitted from 20 WiFi devices \cite{ICCpaper}. 
Apart from the SEI dataset, we also evaluated DLR on our protocol recognition dataset, comprised of RF bursts collected from four ISM emitter
classes: WiFi 802.11n \cite{80211n}, Bluetooth (BT) \cite{BT}, ZigBee \cite{ZBee}
and NRF \cite{NRF}. We utilized this dataset in \cite{SDAE}, which serves as a reference to compare the DLR effectiveness.

Our innovative architecture of split loops helps preserve the H/W reduction while exceeding the SoA accuracy in both applications. 
DLR may be configured to synergistically combine different input transformations with split loops, which offers flexibility for different ML applications and maintains the same outstanding performance. 
By showcasing 2 different datasets, we demonstrate that DLR supports a range of applications by adjusting its parameters and architectural combinations to achieve low SWaP, high accuracy and low latency. 

\begin{figure}[h]
\vspace{-1mm}
\centering
\includegraphics[width=0.49\textwidth]{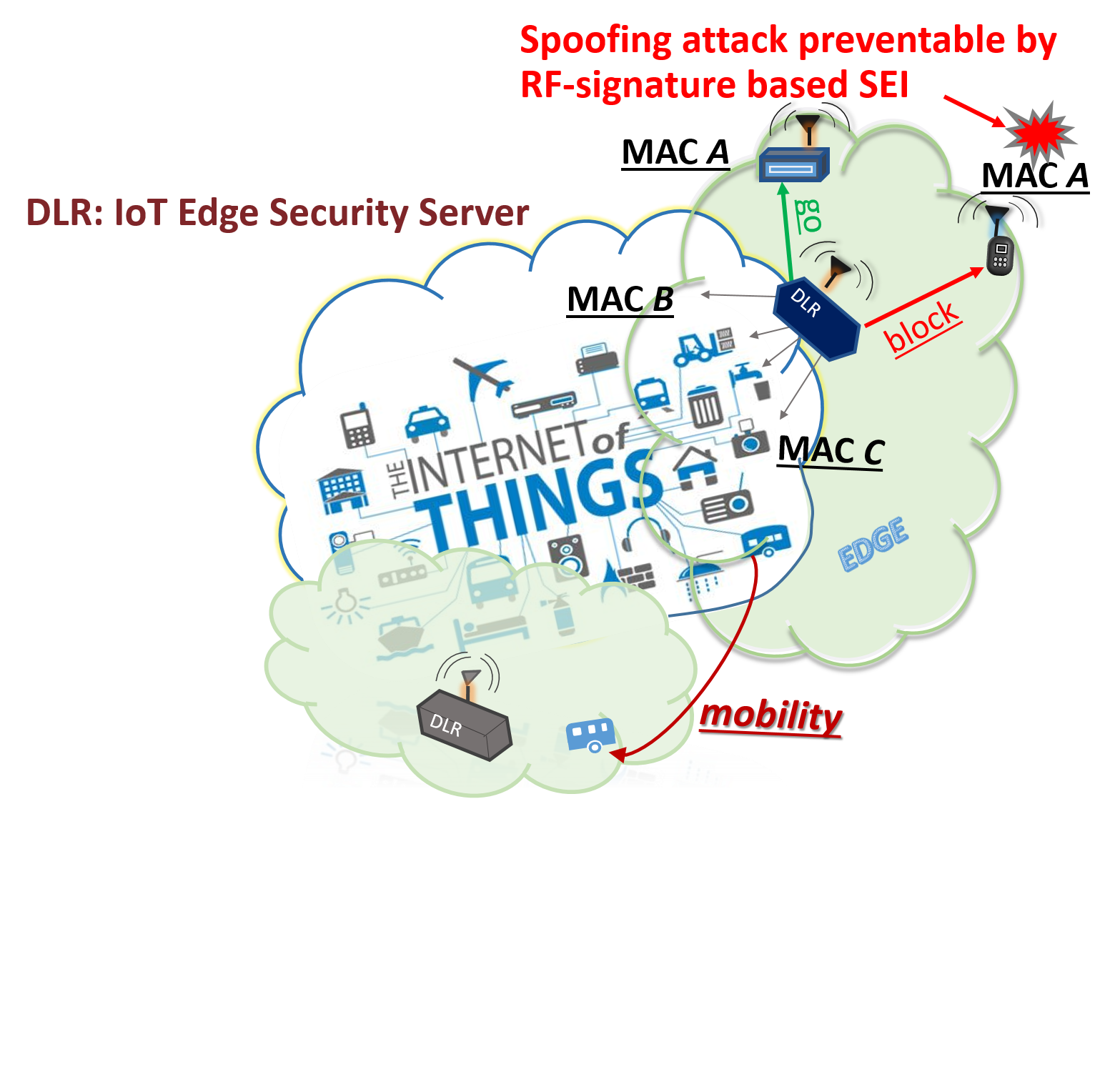}
\vspace{-20mm}
\caption{To authenticate or characterize mobile devices across large regions, individual DLR devices must be first trained on the wireless emissions within range. The trained models can then be exchanged and integrated between DLR systems. The communication cost is proportional to the number of weights $\left\|W\right\|=NT,$ where N is the reservoir size and T the number of transferred devices.}\vspace{-3mm}
\label{fig:dist}
\end{figure}

Finally, to authenticate or characterize mobile devices across large regions, DLR can be trained in a distributed fashion with very little additional processing and a small communication cost, all while maintaining accuracy. Individual DLR devices will be trained on wireless  emissions within range, and the models trained individually will be exchanged among DLR systems, concatenated and possibly re-tuned. This insures that mobile devices that enter the range of a particular DLR will be immediately identified. Please see Fig.~\ref{fig:dist} 
for an illustration.

Although we introduced the DLR in our prior work \cite{ICCpaper}, we did not focus on the edge-based operation with resource constrained devices, nor did we explore distributed operations.
This new part is independent of the H/W implementation of DLR, and concerns the learning to classify subsets of devices or protocols, and then merge the classifiers. 

The proof of concept was done on SEI for both disjoint and overlapping sets of devices, which correspond to the wireless range of DLRs that exchange and retrain individually trained SEI classifiers. We are not aware of any prior work on merging the classifiers based on delay-loop reservoirs in a decentralized framework.

Our new decentralized training architecture supports the following: we train a subset A on $DLR_1$; we train the subset B on $DLR_2$ (assumed to be a remote device, but a collocated DLR would also do).
If $DLR_1$ and $DLR_2$ do not have common devices in their range, then we take their respective weights $W_{out1},$ and $W_{out2}$, transfer it to a single layer neural network with some regularization, and in a couple of epochs of retuning we get $>$ 99\% for 20 devices. This is the absolute highest accuracy we have achieved with any approach. Even without retraining the accuracy is above 98\%, which indicates that distributed training equips a DLR to identify devices that are not in its range yet.
If A and B are overlapping, then the approach is similar but $W_{out1}$ and $W_{out2}$ are transferred in a more elaborate way.

Note that one may also want to transfer a single weight matrix $W_{out}$ to a neural network (i.e., if you train on the entire dataset). This addresses the case when we choose to train efficiently using Ridge Regression, but we want to continuously adapt the trained model to new or modified signatures (e.g., dynamic wireless channel, new DLR receivers). In any case, we will not experience performance loss due to the weight transfer. Alternatively, if we take the same dataset of reservoir states and train the same single-layer network with it, only without transferring the weights, we achieve just about 40\% of accuracy after hundreds of epochs. The advantage of the proposed approach is obvious, and it has a great potential for Edge applications.

The organization of this paper is as follows: Section II describes motivation for the featured applications; our algorithmic and architectural solutions based on the implementation of the delay-loop concept in DLR are described in Section III; Section IV discusses experimental results, including accuracy for different configurations and H/W reduction figures for the SEI platform; Section V describes a neural-net based integration of locally trained DLR classifiers; Section VI discusses distributed learning infrastructure and discusses the cost for particular cases of interest to SEI-based authentication. The detection of devices not covered by a DLR is also discussed there; Section VII concludes. 
\section{Motivating Applications for In-situ Decentralized Training}
5G and open radio access networks (Open RANs) will result in hardware deployments that require additional effort to mitigate wireless security risks. As all electronic devices have fingerprints due to manufacturing variability, so do radio frequency emitters. SEI based on RF fingerprinting can individually identify a multitude of wireless devices \cite{CabricSEIAuthor}. SEI can be particularly useful in IoT with billions of small devices with diverse cyber-security vulnerabilities including authentication and emitter geo-tracking. The SEI authentication is passive, hence efficient, and the fingerprint cannot be emulated, which prevents attacks such as MAC address spoofing. However, the SoA approaches to SEI require big servers for training, and lack in-situ training solutions, hence the need for solutions like DLR \cite{SEIIoTJnl}. Previous attempts at in-situ training required extensive HW-specific pre-processing and have been evaluated only on simulated data \cite{PUF}.

Additionally, neural network (NN) based wireless receivers will soon become a reality. Trained neural networks are replacing various signal processing algorithms deployed in the traditional receivers that are compliant to specific wireless standards (see \cite{deepunfolding,modelbased} and references therein). To make them adaptable, we can use DLR in an intelligent RF spectrum sensor,  to quickly identify the protocol of the incoming signal and feed it into a specifically trained stack of NNs. For example, our wireless protocol recognition (WiPRec) DLR would allow us to realize an AI-defined receiver that can communicate across all protocols utilized in the ISM band. 

%
%


\section{Deep Delay Loop Reservoir (DLR) Algorithms} 
\begin{figure}[h]
\vspace{-1mm}
\centering
\includegraphics[width=0.49\textwidth]{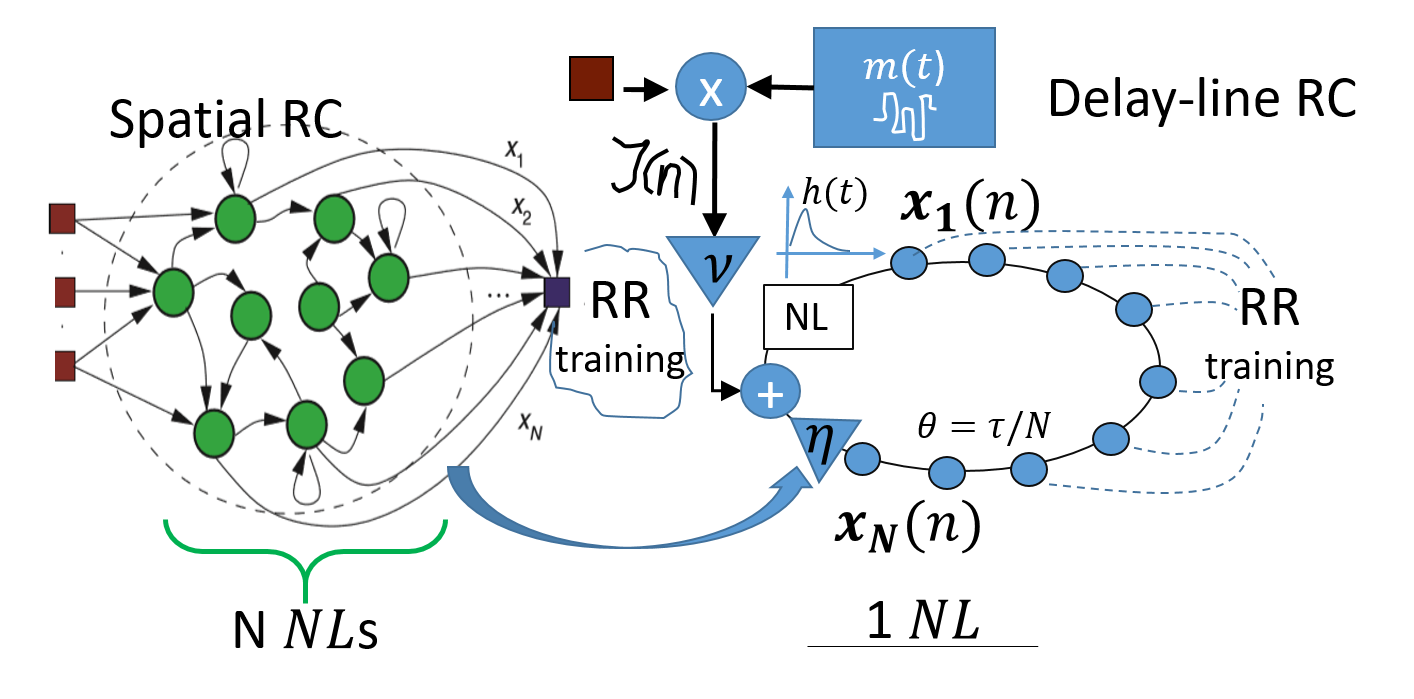}
\vspace{-5mm}
\caption{Delay-loop reservoir replaces the N neurons (NLs) on the left with a single one by using upsampling ($\theta$ depends on the loop bandwidth) and a simple design shown on the right}\vspace{-3mm}
\label{fig:basicdl}
\end{figure}

In its simplest configuration, DLR can be compactly implemented with a single delay loop. In certain cases multiple delay-loop layers, parallel and/or sequential,  may increase the learning capacity of the DLR with no added latency. Each delay loop reservoir in the DLR contains only one non-linearity (neuron), replacing the N neurons in the traditional spatial implementation of the reservoir with N passes through that single neuron. The N-fold increase in delay by the sequential passing of the data through the single neuron (non-linearity NL) is canceled out by an N-fold upsampling (sample and hold) done by a random spreading sequence $m(t)$ (see Fig.~\ref{fig:basicdl}). 
With reservoirs, the N must be large and such high rate upsampling is amenable to a photonic loop implementation. The wide bandwidth and wavelength diversity of photonic loops is the reason why DLR platform also supports photonic DLs as a vehicle to future scalability. The N which allows the reservoir to linearly separate input classes for easier learning may be very large for certain applications, as the classification complexity grows at least quadratically with N. However, there are also other means of reducing the required reservoir size, which we present in this work.
In this paper we report results based on FPGA and software (digital) implementations of the loop, as the emphasis is on algorithmic and architectural solutions. We are also emphasizing the training versus inference, as it is a more demanding aspect of DLR. Once trained the DLR maps the signal to its unique transmitter by just passing the burst of samples through the loop and multiplying it with the weight matrix. For a burst of 1 us duration, inference happens in sub-millisecond including the extraction of the burst and its preprocessing by transforms. Figures of merit (FOMs) that we report for training include the accuracy on the testing set, $H/W$ complexity, memory and latency.

\subsection{Dataset and Preprocessing}
\begin{figure}[h]
\vspace{-1mm}
\centering
\hspace{-1mm}\includegraphics[width=0.5\textwidth]{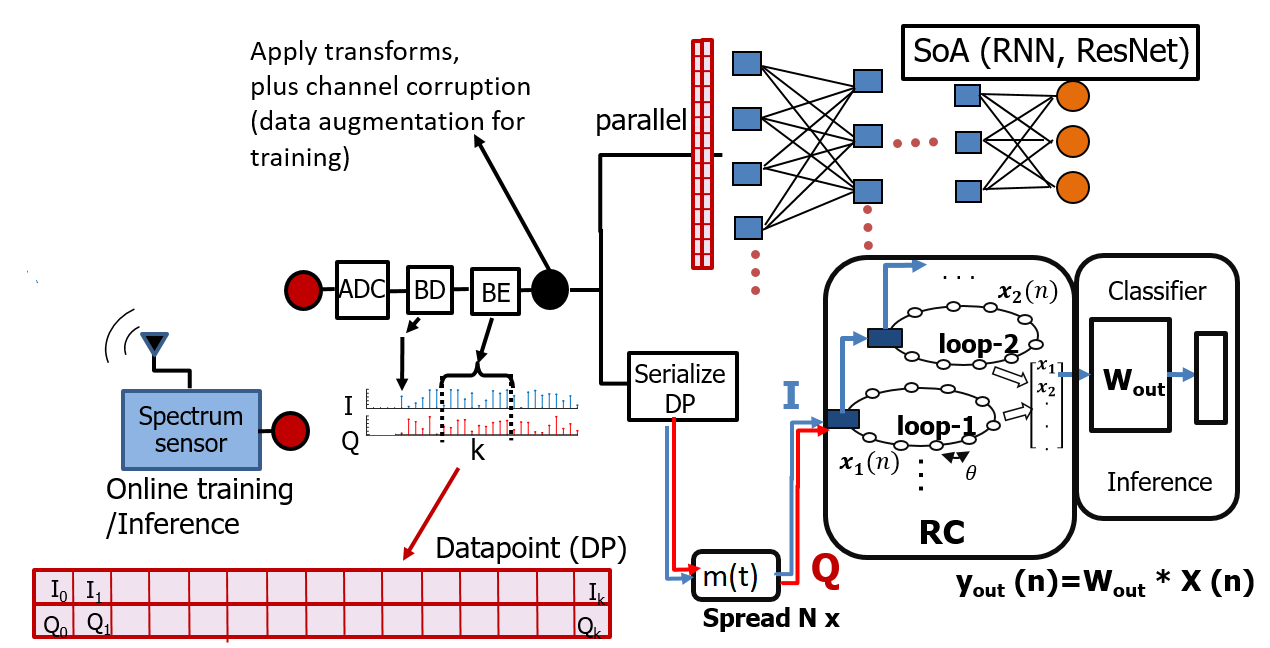}
\vspace{-3mm}
\caption{DLR system for SEI/WiPRec: BD is burst detection, while BE stands for burst extraction. RC denotes delay loops that map the $n$th datapoint into its state vector $X(n).$ $\curls{X(k)}$ are used to train the coefficients $W_{out}$ utilized in the inference $ID_n = argmax\paren{W_{out}X(n)}.$
}\vspace{-3mm}
\label{fig:DLRsys}
\end{figure}

\begin{figure}[h]
\vspace{-1mm}
\centering
\hspace{-1mm}\includegraphics[width=0.3\textwidth]{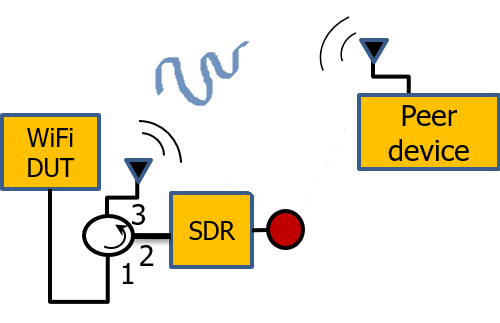}
\vspace{-3mm}
\caption{Controlled capture of high SNR WiFi waveforms.
}\vspace{-1mm}
\label{fig:cabled}
\end{figure}

\begin{figure}[h]
\vspace{-1mm}
\centering
\hspace{-1mm}\includegraphics[width=0.5\textwidth]{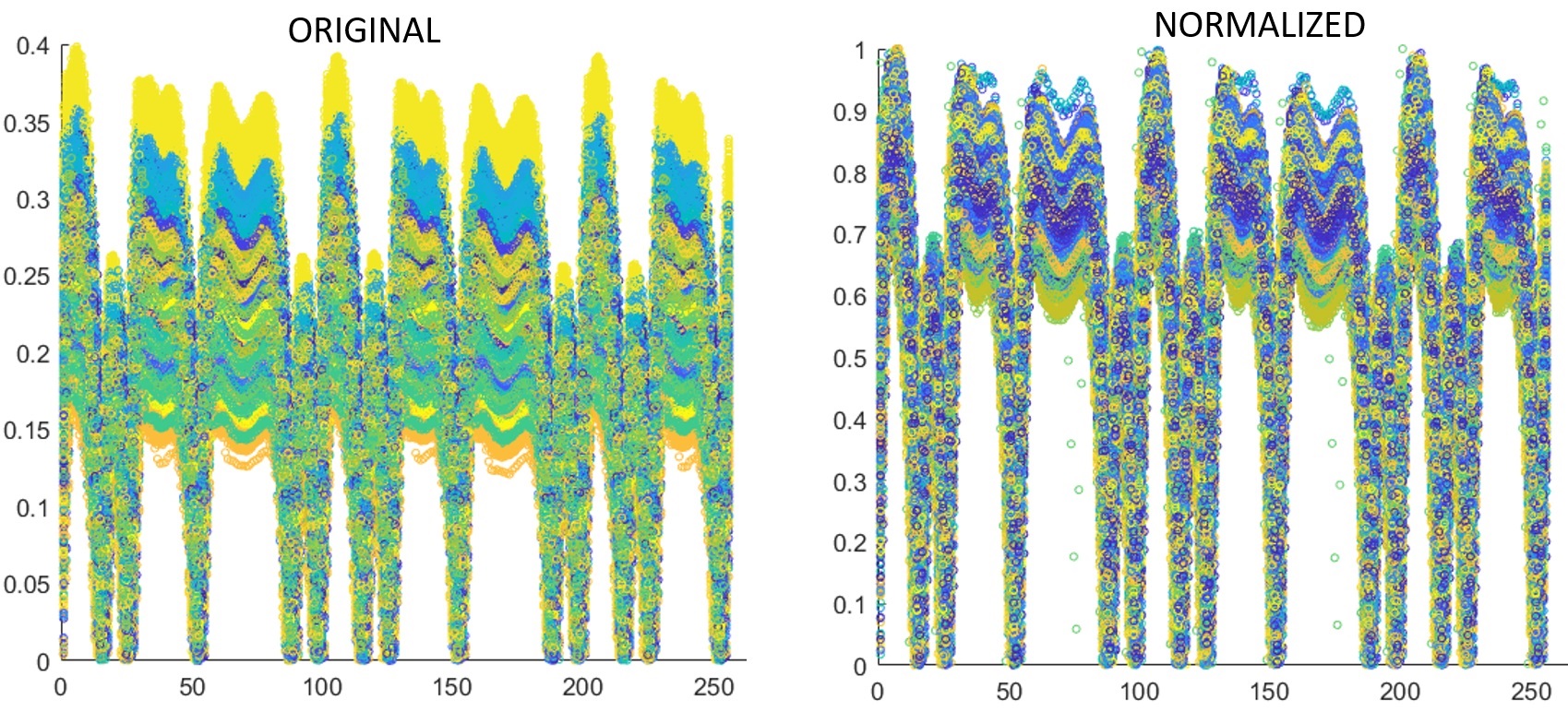}
\vspace{-3mm}
\caption{Power level (y axis) is not a salient feature for SEI and must be removed by normalization. All test datapoints (of length 256 samples each - x axis) have been color-coded according to their specific emitter and scatter-plotted.
}\vspace{-3mm}
\label{fig:normalization}
\end{figure}

The dataset to train the SEI detector contains 20 classes, corresponding to 20 distinct WiFi devices. We first started with the dataset of 4 devices but it was not  sufficiently complex to allow for the exploration of the techniques that scale the design, such as split loops. To build the dataset we captured the emissions of commercial WiFi devices as they were sending access requests to an access point while using the same spoofed MAC address. 
The emissions were captured by a USRP X310 with UBX RF daughterboard, with a sample rate of 100 MHz centered in the middle of the 2.4 GHz ISM band. SEI datapoints are created by extracting bursts of 1024 complex (I/Q) samples from the captured time-series, counting from the sample immediately after the detection of the rising edge of the signal. For the ISM application, bursts are extracted from random places within the active emission capture. Further preprocessing steps are explained in the subsection on transforms. Please see Fig.~\ref{fig:DLRsys} which describes the information flow and the process performed by the DLR platform in Fig.~\ref{fig:platform}. The red circle marks the input into the ADC, which performs digital sampling of the captured waveform, and matches the red circle in Fig.~\ref{fig:platform}. Training data was captured using a cable connected to the antenna via a circulator and deposited to an on-board storage withing the DLR platform (as shown in Fig.~\ref{fig:platform} at the end of the yellow path). Said capture, depicted in Fig.~\ref{fig:cabled}, provides high SNR datapoints to which we may add noise and channel corruption (black circle) in a controlled way using our policy-based data augmentation system \cite{policybased}.  
We trained our SEI DLR system using 600 bursts (datapoints) per device. Note that we normalize all datapoints before spreading them by the loop mask, to ensure that only salient features are used for classification. Power levels cannot be used as signatures of different emitters because the propagation channels attenuate the signal power depending on the receiver, and the emitting power can also vary arbitrarily for the same device. Fig.~\ref{fig:normalization} illustrates 
the difference between magnitudes of normalized and unnormalized datapoints (of 256 samples each): for unnormalized data the power-based differences between classes are discernable even visually since we color-coded datapoints according to their class (emitter). This caused an illegitimate increase in accuracy when unnormalized data was used for the preliminary results in \cite{GomacTech}.

\subsection{Delay Loop (DL)}
The basic algorithm for the delay-loop state is expressed by~\eqnref{delayloop}.  
\begin{align}
\nonumber &X_k (n)=\\
&\int_{\delta_k-\tau-\epsilon}^{\delta_k}{h_{\delta_k-\tau-\delta} f_{NL} \set{\eta X_\delta (n-1)+\nu J(n)(\delta + \tau)}d\delta.}  \eqnlabel{delayloop}
\end{align}
 
\begin{figure}[h]
\vspace{-1mm}
\centering
\includegraphics[width=0.39\textwidth]{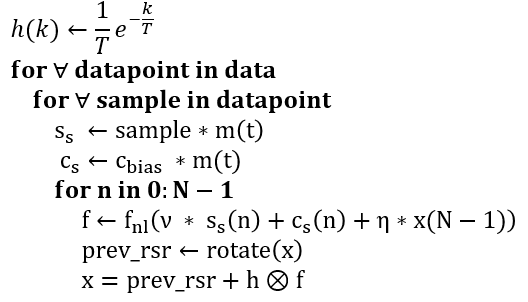}
\vspace{-4mm}
\caption{Pseudo code for the delay loop}\vspace{-3mm}
\label{fig:pseudo}
\end{figure}

$X_k$ is the $k^{th}$ virtual element of the state vector $X.$ The upsampled time is defined in chips $\theta$ (in Fig.~\ref{fig:basicdl}), which gives rise to $J(n)(t),$ also denoted as $s_s(n)$ in the pseudo-code, Fig.~\ref{fig:pseudo}.  Each sample $s(n), n\in{1,\cdots,\ell}$ of the input datapoint of size $\ell$ (Fig.~\ref{fig:DLRsys}) is spread by the mask $m(t)$ and clocked into the loop as $J(n)(t),$ chip-by-chip. Here, $t \in {1,\cdots,N}$ is the chip-time index, and $k \in {1,\cdots,N}$ is the loop position index. Note that $N$ is the number of virtual reservoir nodes, as well as the length (in chips) of the spreading sequence (mask) $m(t)$ as shown in Fig.~\ref{fig:basicdl}.  $J(n),$ also indicated in Fig.~\ref{fig:platform}, is the input to the loop, and the output is read out after the last of the $\ell$ samples is clocked-in and put through the loop's non-linearity $N$ times.  

Each chip of the spread sample $J(n)(t)$ is put through the nonlinearity $NL$ as a linear combination with the tail of $X: X_N(n-1)$   (which has been affected by the same non-linearity at time $t-\tau,$ i.e., by the previous input sample $n-1$, where $\tau = N\theta$). Summation of the spread data input and  the tail of $X$ at every $t$ is practically creating the edges of the recurrent layer from the spatial implementation of the reservoir (the left side of Fig.~\ref{fig:basicdl})  \cite{AdvancesinphotonicRC}.  The randomness of edge weights is determined by the randomness of the spread sequence, which unfolds the edges at chip time. The output of the $NL$ may be convolved with filter $h(t)$ (Fig.~\ref{fig:pseudo}) to model the $NL's$ temporal response in photonics \cite{PhotRCTut},  which occurs on a finer time scale $d\delta$.  $\delta_k$ in the integral is the current time at node $k.$ 

We refer to the $X_k$ as the virtual node. No extra function or transformation is carried out on $X_k$ after the filtered non-linearity $f_{NL}.$ Omitting the propagation details, $X_k$s are simply time-shifted, in chip time $\theta,$ which also roughly matches the propagation time in photonics. \eqnref{delayloopD} models the shifting of the $f_{NL}$ output in digital implementaion, where $X_k$ at chip time $t$ is given by 
\begin{align}
\nonumber &X_k (t)=\\
 &\sum_{u=0}^{1}{h(u) f_{NL} \set{\eta X_k(t-N +u)+\nu J(t-k-u)}}  + \sigma,\  \sigma \longrightarrow 0. 
\eqnlabel{delayloopD}
\end{align}
The loop gain parameter $\eta$  and input gain $\nu$ must be calibrated to provide a proper dynamic state of the reservoir. $h(t)$ in \eqnref{delayloopD} is sampled at chip time.  
The results are based on $f_{NL}=sin\paren{\cdot},$  although we also used $tanh\paren{\cdot}$ as the neuron, with similar effects. The pseudo-code in Fig.~\ref{fig:pseudo} is mappable to the photonic implementation, which was important for the development and calibration of the photonic loop allowing the comparison with the baselines achieved with the digital-only loop \cite{PhotRCTut}. 

Another innovation is the introduction of “deep”, multi-layer architectures with both stacked (sequential) and parallel (split) reservoir loops.  The rationale for the latter is that splitting the input into $k$ parallel loops and combining their outputs reduces the complexity of the classifier by $k^2$ (please see section~\ref{subsec:split}). For the former, we explored if $s$ sequential loops increase the representation complexity of the input allowing for more accurate classification. As the accuracy achieved to date by using sequential loops has not exceeded the one obtained with split loops only, and for space considerations, we here discuss just the split loops. We use transforms to adapt the salient signal information to the split loop input. We develop multiple ways to merge the outputs of the constituent DLR loops into a single output to be used to train a simple algorithm (Ridge Regression \cite{RR}). 

We used Bayesian hyper-parameter optimization to find the best parameters, including the splitting. The idea is to treat the RC as a “black-box”, computationally heavy function: $f(\nu,\eta,N,s,k,\cdots)\rightarrow\set{0, 1},$ where the output is the resulting predictive accuracy.
Succinctly, the process involves the application of a Gaussian process prior on $f()$, then as more sets of parameters are sampled, the prior along with these parameters forms a posterior distribution over $f$.
We combined this with a hierarchical grid search  of the hyper-parameter space (especially for the composition of loops) to determine values of the optimal parameters, given a particular input transform applied to datapoints. This is done only once per application. It should be emphasized here that the loops are never trained, only the RR algorithm which uses their outputs.

Note that the computational complexity of training the DLR is the complexity of the last stage (Ridge Regression), given the simple implementation of the loop.
\subsection{Ridge Regression}
The Ridge Regression (RR) model for the estimation of the weight coefficient matrix $W$ is calculated as:
\begin{align}
W=argmin_{W_{out}}\paren{\sum_{j=1}^{B}{\left\|y_j-W_{out}X\right\|_2^2+\lambda\left\|W_{out}X\right\|_2^2}},\eqnlabel{RR}
\end{align}
where $B$ is number of training datapoints,  $y_j\in Y_{out}$ is the one-hot device-label corresponding to the $j^{th}$ training datapoint, $W_{out}$ is the output weights to be trained, and $\lambda$ is the regularization factor. The RR equation can also be written in the closed form  $W = (X^T X +\lambda I_N )^{-1}  \ (X^T Y_{out} )$, where $X$ is the matrix of the B state vectors used for its training. 
\begin{figure}[h]
\vspace{-1mm}
\centering
\includegraphics[width=0.49\textwidth]{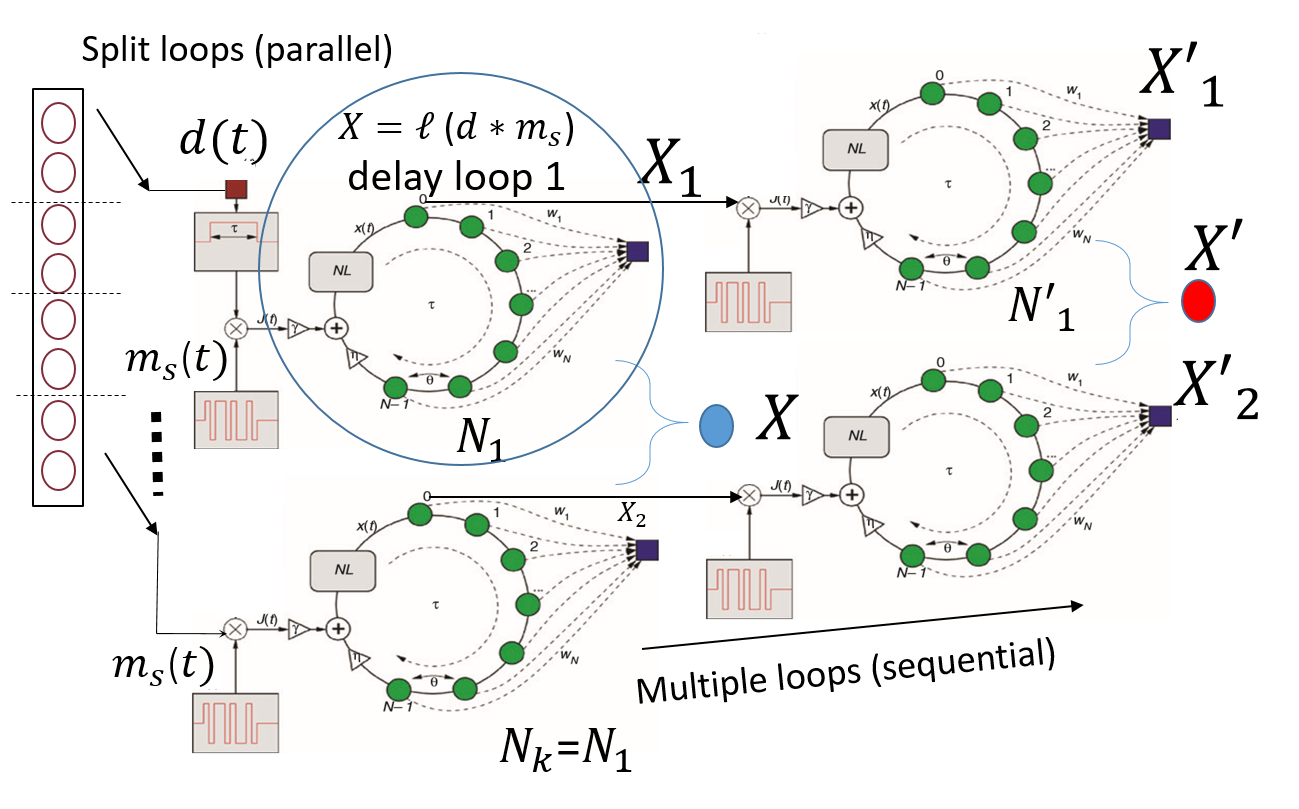}
\vspace{-6mm}
\caption{Using k split (parallel loops) reduces the size of the spreading mask and, hence, the size of each split loop while achieving the required projection into higher dimensional space. The joint state vector $X$ (marked by the blue circle) can be obtained as $\sum_{j=1}^{k}{X_j}$, or as a normalized scalar product of $X_j$s.}\vspace{-3mm}
\label{fig:multiple}
\end{figure}
\subsection{Split Loops}\label{subsec:split}
As Fig.~\ref{fig:multiple} shows, the split loops process the $k$ disjoint pieces of the split datapoint in parallel and result in the k state vectors $\curls{X_1,\cdots, X_k}$. We can pass these $k$ outputs through another layer of parallel loops. The joint state vector $X$ (marked by the blue circle in Fig.~\ref{fig:multiple}, or red circle for a 2-layer design) can be obtained as $\sum_{j=1}^{k}{X_j}$, or as a normalized scalar product of $X_j$s. Both gave similar results in terms of the improved accuracy over a single loop. Let us emphasize here that DL is a dynamic system, and the reservoir size $N$ for a datapoint of size $\ell$ must be large enough to bring the DL into a dynamic state where the class separation happens.
Each of the split-datapoints is of size $\ell/k$ and its $N_j<N, j=1,\cdots,k$ now may be $k$ times smaller than the $N$ without splitting. Now, as the total $X$ is a linear combination (and not concatenation) of the split loops' $X_j$s, the entry to the RR is still of size $N_j$, which reduces the RR complexity from $BN^2$ to  $B(N/k)^2$. Beyond certain value of $k,$ the accuracy starts to drop as splitting the datapoint affects the samples that are no longer independent. The exact value of the threshold $k$ depends on the input transform applied to the burst of $I/Q$ samples composing the dataset. As an example, this threshold for the FFT is above 10 for the burst length of 1024 samples.
  
The joint state vector results in higher accuracy than any of the $k$ split state vectors taken separately. For now it is important that the results obtained using the addition and scalar multiplication exceed the SoA, but we would also like to quantify the information loss if any (due to Data Processing Inequality). If we concatenated the split state vectors we would have preserved the information from the split datapoints completely but the resulting length of the state vector would have increased the complexity of learning by $k^2$. An information-theoretic analysis is under way on how much information is passed from the input to the state vector under different operations to construct the joint state.  
\subsection{Input Transforms}
It is known that certain transforms may losslessly compress the data if the information content is sparse, and some are more robust to noise. Motivated by this, we experimented with multiple input transforms in order to best mach the input to a specific architecture of DLR, given constraints in the complexity (i.e., largest reservoir output we are willing to train). 
Since the RF waveforms are characterized by amplitude and phase we have 2 dimensions per sample in each 1024-long datapoint. Note that DLR loops cannot process complex-valued datapoints directly, and therefore all the transforms that we apply lose the phase information (except for the case in Fig.~\ref{fig:mix}) 
\subsubsection{Complex Amplitudes}
\begin{figure}[h]
\vspace{-1mm}
\centering
\includegraphics[width=0.45\textwidth]{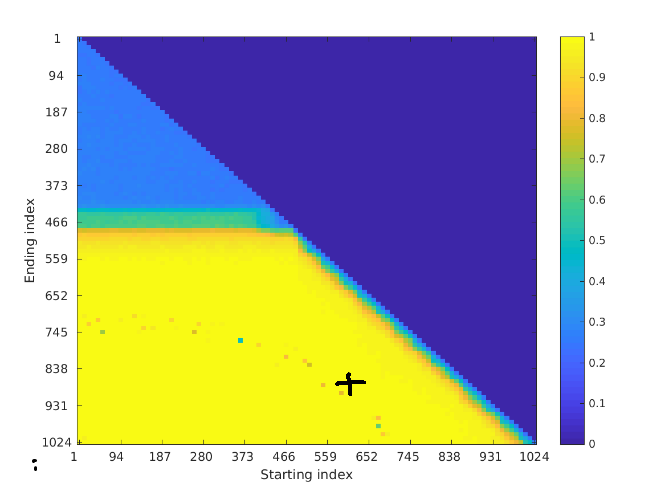}
\vspace{-5mm}
\caption{The salience map across the datapoint of 1024 samples of normalized complex magnitudes; the best subburst of 256 samples is marked by an $\times$.}\vspace{-3mm}
\label{fig:salmap}
\end{figure}

We consider the transform from complex-valued samples to their amplitudes as the baseline. Here, subbursts of 256 complex amplitudes are extracted to replace the longer datapoints of 1024 samples. For this extraction we used a simple linear classifier to explore how salient information is distributed along the datapoint. We conducted a systematic experiment in which we have been replacing each datapoint with a subset of contiguous complex-valued samples within that datapoint, and evaluating the accuracy of the classifier for different sizes of such subsets, and different offsets from the start. Please see Fig.~\ref{fig:salmap} which shows color-coded accuracy values for all combinations of the starting (x-axis) and the ending edge (y-axis) of the burst within the initial 1024 samples (including the one that takes all samples). The best position, marked by an $\times$, results in the shortest among the subbursts that achieve the highest accuracy. The rationale was that if the salient information was concentrated in a smaller sub-burst, than the size of the reservoir $N$ can also be scaled down while maintaining the same accuracy.  The 256 samples per datapoint, extracted at a location that preserves the accuracy achieved with the original burst, are equivalent to a 1us long burst of samples. 

\begin{figure}[h]
\vspace{-1mm}
\centering
\includegraphics[width=0.37\textwidth]{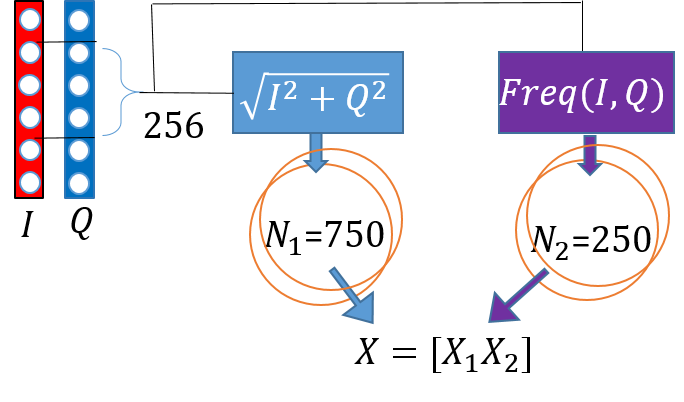}
\vspace{-5mm}
\caption{Using diverse transforms showed some initial promise in combining complementary transforms - magnitudes are combined with simple frequency estimates}\vspace{-3mm}
\label{fig:mix}
\end{figure}

The accuracy of DLR based on the amplitudes of the subburst of 256 complex samples ranks the worst among the tested transforms. Please observe how the accuracy for 2 split loops in Fig.~\ref{fig:acc} gets higher with the reservoir size when amplitudes (blue - plain vanilla transform) are replaced by the amplitudes of complex FFT transforms. We have also combined the amplitudes in one pair of split loops with the frequency estimates in another and concatenated the state vectors. With the loops processing the magnitudes of dimension 750, and the loops processing the frequency estimates of dimension 250 (as in Fig.~\ref{fig:mix}),  we get the total state vector of length N=1000 and the accuracy of 93.8\%. Note that frequency estimates \cite{SKay} of the complex valued input bursts are real valued bursts of smaller length than the original bursts, as we are using 3 consecutive I/Q samples to calculate one frequency value. This approach should be further explored with a finer frequency estimate, and exhaustive loop combinations.
\subsubsection{FFT}
The exploration of the information salience above revealed the sparsity of the original dataset that can be leveraged through additional transforms. We next optimized the performance of the digital test bench when the input to the loop are magnitudes of the FFT transform applied to the bursts of complex valued RF samples. The initial findings were that the FOMs deteriorate since the FFT transform ‘spreads’ the information across the datapoint. Specifically, without the FFT, the salient information is limited to the subburst of length 256 located in the middle of the 1024 long burst of sampled signal magnitudes. When the FFT magnitudes are used as datapoints, the salient information is located in all 1024 samples. This requires a delay loop longer than 1024, exceeding the loop size of 600 or 800 which produced the best accuracy-complexity trade-offs.
\begin{figure}[h]
\centering
\hspace{-4mm}\includegraphics[width=0.52\textwidth]{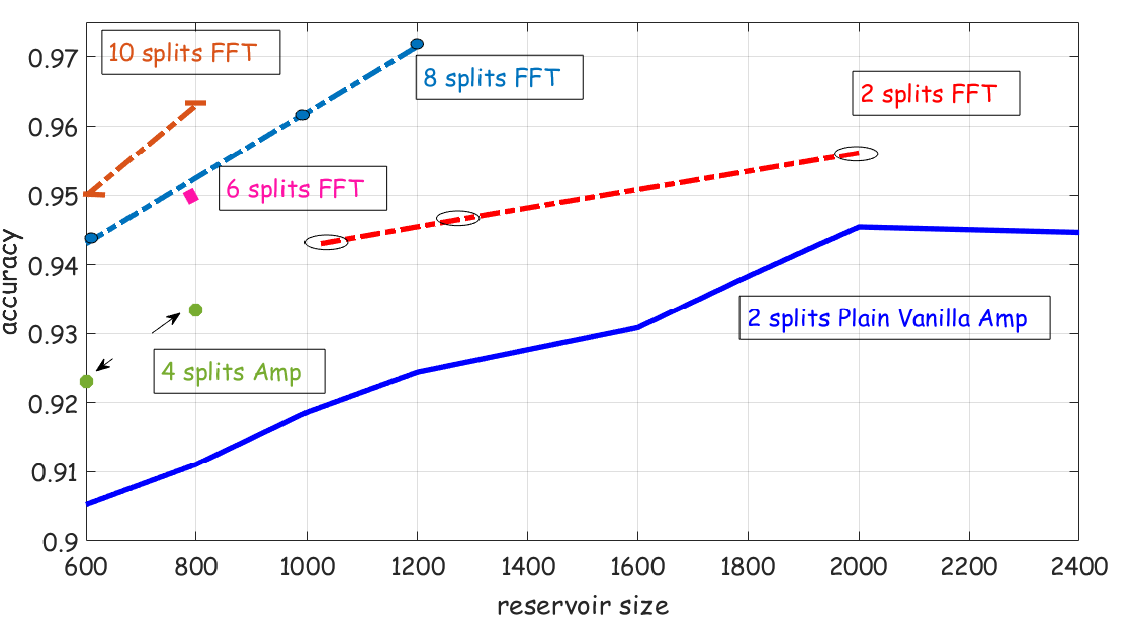}
\vspace{-3mm}
\caption{Accuracy as a function of the reservoir size for split loops when all are using the same input transforms - either signal amplitudes and FFT amplitudes. Notice the improvements with 10 splits, with unchanged complexity.
}\vspace{-4mm}
\label{fig:acc}
\end{figure}
However, using the technique of loop splitting, we were able to utilize 8-fold splits to decrease the input dimension, and consequently reduced the reservoir size in each of the loops  to the previously used values. While this kept the H/W reduction factors unchanged, we achieved a gain in accuracy. Please see Fig.~\ref{fig:acc}. Note that splitting the datapoints that are based on the complex signal amplitudes into 8 loops deteriorates the accuracy compared to 4 loops, which is omitted from Fig.~\ref{fig:acc} due to clarity. This makes the FFT more effective in achieving the best accuracy-complexity trade-off.
\subsubsection{Differential FFT}
The approach here is to find the average of the burst amplitudes across the training dataset and remove it from each datapoint amplitude, then create new complex-valued bursts with  the common bias so removed (while preserving the phase). We perform the FFT of the resulting waveform, referring to its magnitudes as the {\em differential FFT} transform. This transform helped increase the robustness of the photonic loop by reducing its sensitivity to the in-loop noise: compared to the regular FFT magnitudes, differential FFT increased the accuracy by more than 15\% for the same reservoir size. In the digital loop, the differential FFT did not make a difference.
\subsubsection{Decimated DFT}   
\begin{figure}[h]
\centering
\hspace{-2mm}\includegraphics[width=0.5\textwidth]{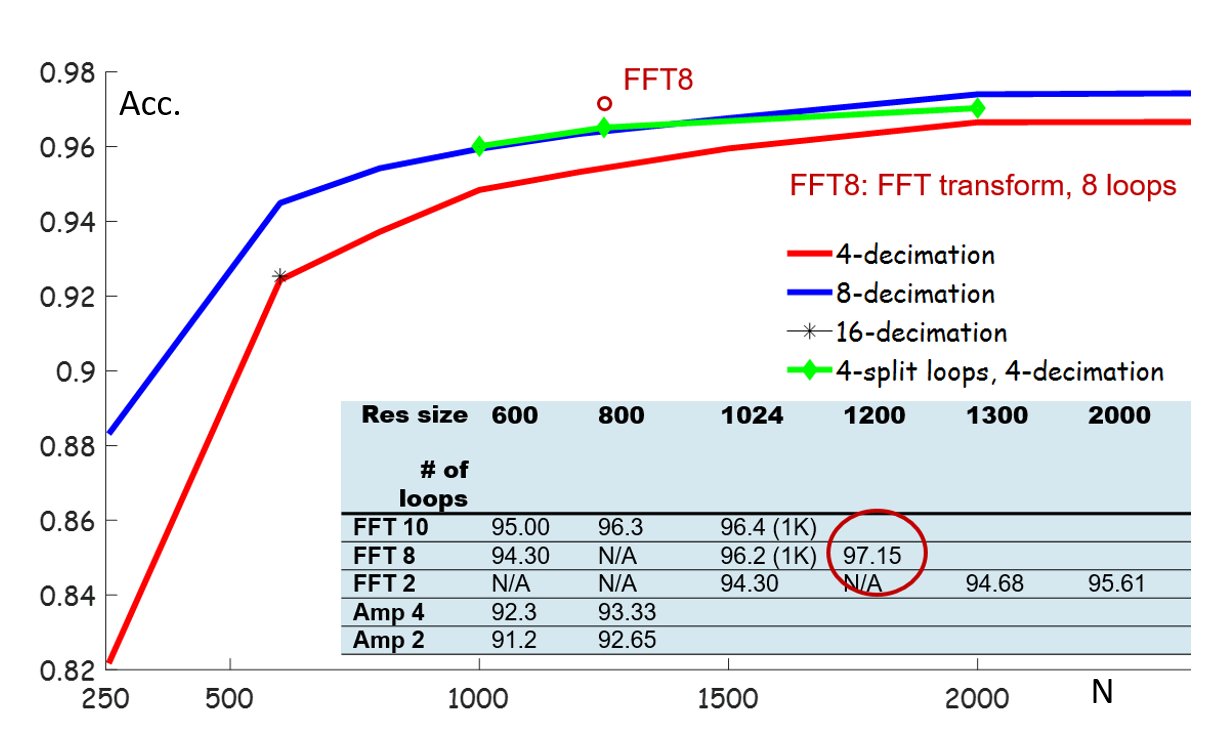}
\vspace{-3mm}
\caption{Accuracy vs. N with decimation of DFT: N=1200 achieves 97.48\% of accuracy with 2 split loops and $d=8,$ or 4 split loops and $d=4.$
With further decimation (16-fold) the accuracy falls back to 4-fold decimation (2 loops).
}\vspace{-5mm}
\label{fig:accDFT}
\end{figure}

We decimate the DFT matrix 
\begin{equation}   \label{eq:DFT}
    D = \frac{1}{N}
    \left( \begin{array}{ccccc}
        1 &  1 & 1 & \ldots & 1 \\
        1 & \omega_N^{-1} & \omega_N^{-2} & \ldots & \omega_N^{-(N-1)} \\
        1 & \omega_N^{-2} & \omega_N^{-4} & \ldots & \omega_N^{-2(N-1)} \\
        \vdots&\vdots&\vdots& &\vdots\\
        1 & \omega_N^{-(N-1)} & \omega_N^{-2(N-1)} &\ldots&\omega_N^{-(N-1)^2} 
    \end{array} \right),
\end{equation}
where $\omega_N = e^{i 2\pi/N} \in \C$, by some factor $d$ - by keeping every $d^{th}$ column of $D$ - creating a sparse DFT matrix $D_d$ of size $1024 \times 1024/d.$ By multiplying the complex-valued burst of size 1024 by $D_d$ we obtain a decimated discrete frequency transform whose length is shortened $d$ times. We use its amplitudes as datapoint to split into 2 parallel loops. Please see Fig.~\ref{fig:accDFT} where the accuracy is shown for different  reservoir sizes.  If we use more split loops the decimation factor must be reduced to maintain the accuracy. Given a compressive transform, we observe that the compression affects the number of split loops needed to achieve accuracy gain for the same reservoir size. Split loops become useless after a certain point, which depends on the input transform.
Given this trade-off, an interesting future goal is to compare the cost of performing a per-datapoint transform vs splitting the loop.
\section{Efficiency of Individual DLRs: Experimental Results}
\begin{figure}[h]
\vspace{-1mm}
\centering
\hspace{-1mm}\includegraphics[width=0.5\textwidth]{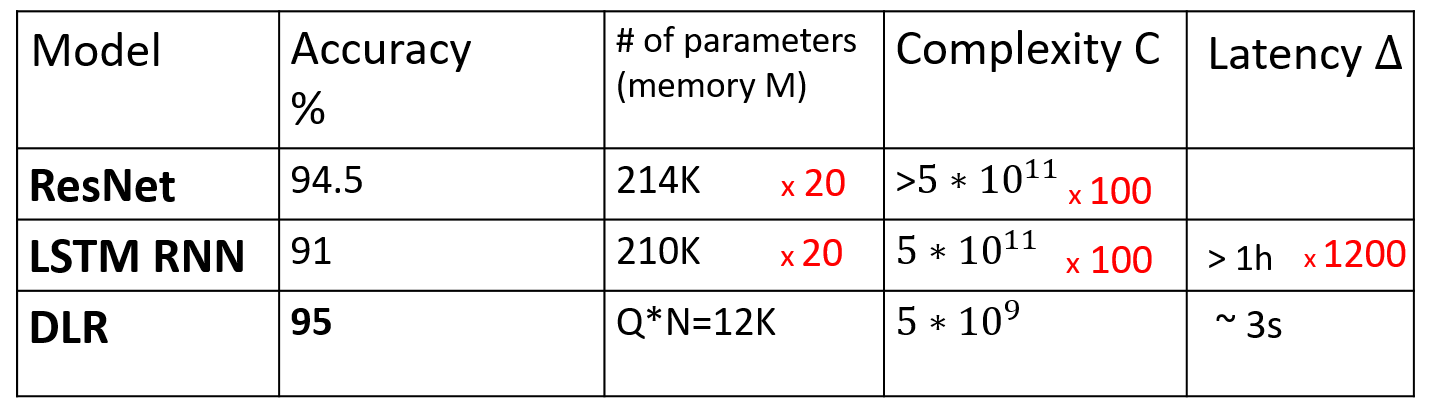}
\vspace{-3mm}
\caption{Figures of Merit compare the best trade-off of {em SEI accuracy vs H/W reduction} for DLR in comparison with SoA. Complexity is expressed in terms of total MAC operations and memory in terms of the number of parameters to train.
}\vspace{-2mm}
\label{fig:fom}
\end{figure}

The table of figures of merit (FOMs) in Fig.~\ref{fig:fom} shows the 20-device SEI performance numbers for the best trade-off we achieved with DLR. These are compared with the equivalents for the SoA neural networks trained on a single GPU. Note that the actual SoA complexity and memory figures are much higher as we reported the least complex networks that may effectively learn given very long training. The red numbers show the reduction factor obtained with DLR compared to the respective values for ResNet or RNN. The spatial reduction factor for DLR vs SoA is the ratio of the number of trainable parameters (20), and the power reduction factor is the ratio of the training complexity figures (100), while the latency compares the hours of training versus 3 seconds achieved with our demo platform ($\geq 1200$). Memory $M$ in Fig.~\ref{fig:fom} is determined by the matrix of weights $W_{out},$ of size $Q\times N$, where $Q$ is the number of classes, e.g., the number of devices covered by SEI. Complexity $C$ is determined by the matrix of RC states $X$, of size $B\times N$, whose pseudo-inverse has complexity $BN^2$ \cite{pseudoinverse}. For details on how FOMs were calculated, please see our paper on intermediate results \cite{GomacTech}.The result here use the reservoir size of 600. Note that we could achieve higher accuracy  of above 97\% by using the reservoir size of 1200, which increases the complexity 4 x (consult Fig.~\ref{fig:acc}).

Finally, we would like to present the results obtained using the same loop hyper-parameters (without retuning) for the application of wireless protocol recognition (WiPRec). We here achieved 99\% accuracy for the classification of the 4 ISM protocols trained on clean signal collections. We collected data from 5 devices per protocol, 4 from a common manufacturer, the 5th from a different one. We either used connectorized devices in conjunction with cables,
circulators and attenuators (as in Fig.~\ref{fig:cabled} or an RF shield box, to ensure pristine (clean) RF data. We used the same USRP and the same sample rate as for the SEI dataset. After the signals were basebanded and the bandwidth information removed by appropriate decimation, this accuracy dropped by about 2 \%. Removing BW information emulates the efforts to hide the identity by transmitting on a different frequency and performing rate adaptation. 

\begin{figure}[h]
\vspace{-1mm}
\centering
\hspace{-4mm}\includegraphics[width=0.54\textwidth]{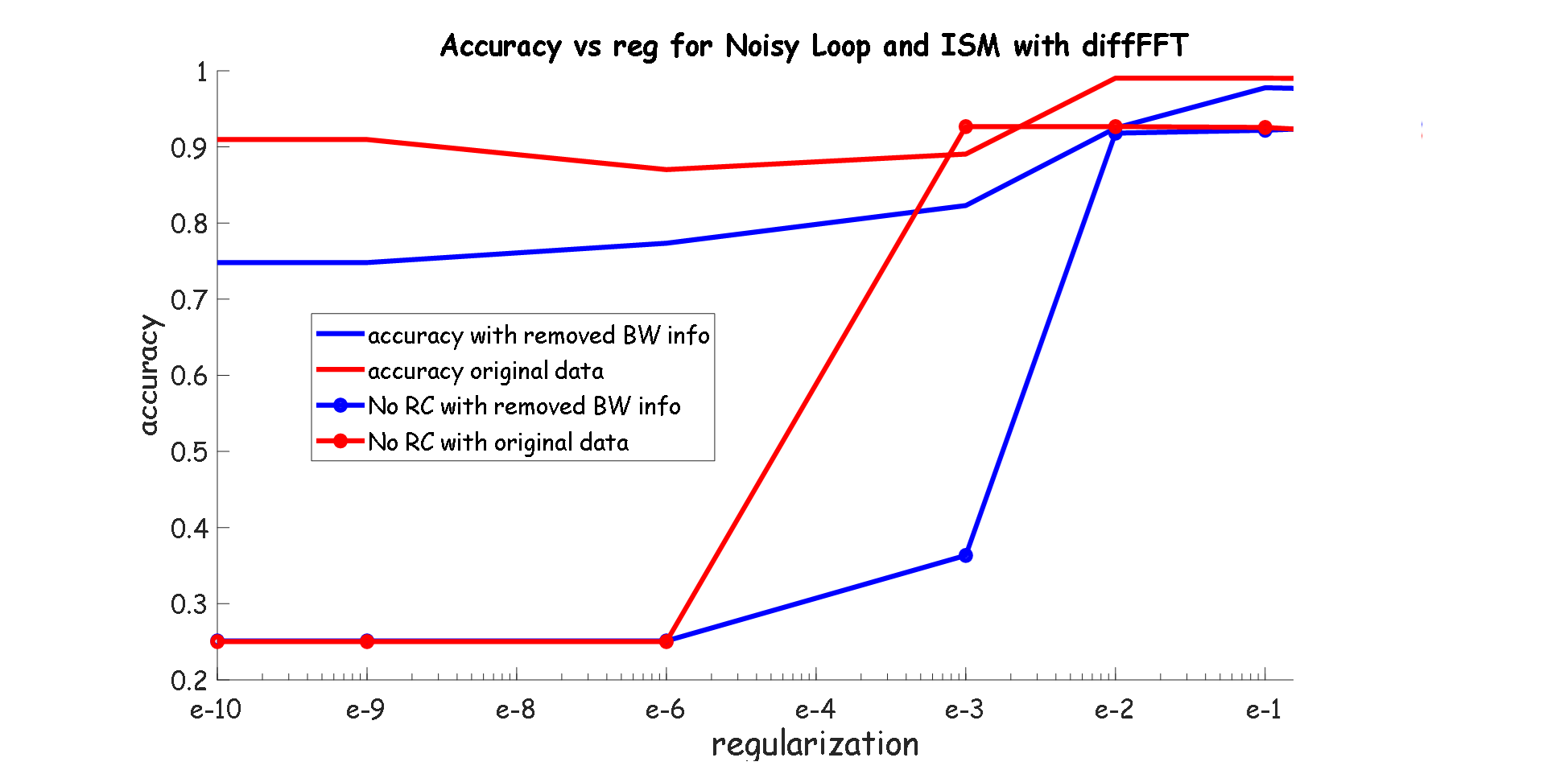}
\vspace{-3mm}
\caption{Accuracy vs RR regularization for classification of the 4 ISM protocols with and without BW as a salient feature.
}\vspace{-3mm}
\label{fig:ism}
\end{figure}

Fig.~\ref{fig:ism} shows both cases on DLR (plots without markers), as well on the simple Ridge Regression (without reservoirs) for the same complexity. The simple RR has a drop in accuracy of about 9\%.  An interesting observation is that the results stayed the same after the loop noise has been added, which was modeled based on the photonic loop observations. We confirmed this result (simulated in the digitally implemented loop) by performing the classification in our photonic loop. DLR also produces a better conditioned weight matrix, i.e. the results are more stable for different values of the regularization parameter $\lambda$ in \eqnref{RR}  than the RR without the reservoir (see Fig.~\ref{fig:ism}). The results  with other transforms are consistent. The idea here was to test if DLR can be re-purposed for other signals, and the results exceeded our expectations as we did not have to re-optimize the parameters. Admittedly, both the ISM dataset and the 4-way classification problem based on it are much less complex. However, comparing it with our published worked \cite{SDAE} that trained an ISM classifier off-line in order to minimize the complexity of inference, we ended up with the 4-times smaller inference complexity, and evidently much simpler and faster training. Future work should include signal corruptions by fading channels and interference.
\section{Neural Net Based Integration of Decentralized DLRs}
\begin{figure}[h]
\vspace{-1mm}
\centering
\hspace{-4mm}\includegraphics[width=0.42\textwidth]{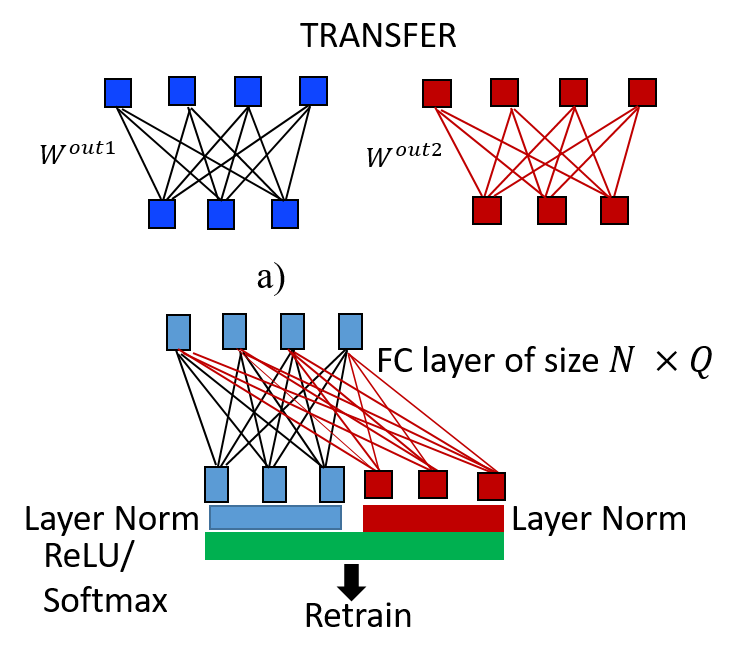}
\vspace{-3mm}
\caption{ Disjoint DLR Integration
}\vspace{-3mm}
\label{fig:distdisj}
\end{figure}

\begin{figure}[h]
\vspace{-1mm}
\centering
\hspace{-4mm}\includegraphics[width=0.38\textwidth]{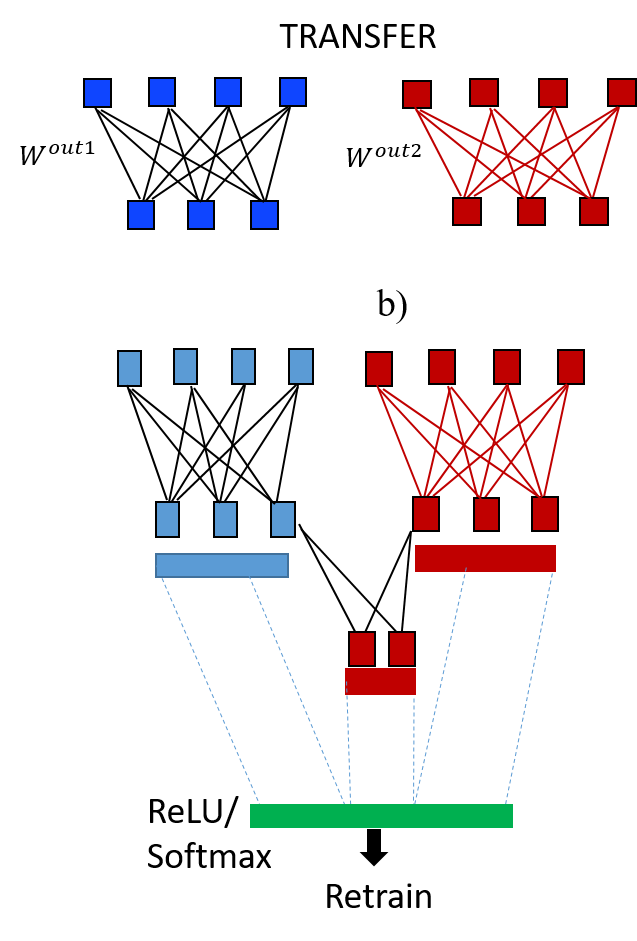}
\vspace{-3mm}
\caption{Overlapping DLR Integration
}\vspace{-3mm}
\label{fig:distovlp}
\end{figure}

Locally trained DLRs can be integrated into a global classifier by means of a neural net composed of a single fully connected layer (a perceptron) with some additional regularization. Depending on whether the subsets of devices covered by individual DLRs are disjoint or overlapping,  we transfer the weights trained in local DLRs to the fully connected layer using one of the two basic patterns presented in Figures~\ref{fig:distdisj}~and~\ref{fig:distovlp}. The idea here is that such a neural net capable of classifying devices globally exists on each local platform (where a locally trained DLR exist).
We will illustrate the two transfer methods ($a$ and $b$) by using 2 DLRs with disjoint or overlapping classes (devices in SEI). For more than two DLRs, the concept is the same. With disjoint sets of labels, we have 
\begin{align}
\nonumber Y_1=\curls{1, \cdots, Q/2},\\
Y_2 \in \curls{Q/2+1,\cdots,Q}.\eqnlabel{methoda}
\end{align}
The respective locally trained weights are obtained using the Ridge Regression fitting function $f_{RR}\paren{\cdot}:$
\begin{align}
\nonumber W_{out1} = f_{RR}\paren{\hat{X_1};Y_1,\lambda},\\
W_{out2} = f_{RR}\paren{\hat{X_2};Y_2,\lambda}, \eqnlabel{locals}
\end{align}  
where  $\hat{X_i},\  i \in\curls{1,2}$ are the reservoir state vectors obtained from disjoint sets of training datapoints matching the sets of labels $Y_i,\ i \in \curls{1,2}$ in locally trained DLRs.
We combine $W_{out1}$ and $W_{out2}$ by transferring them to the weights of the two fully connected (FC) layers according to the color-coded pattern in Fig.~\ref{fig:distdisj}. Let us refer to the neural net in Fig.~\ref{fig:distdisj} as $Net_a(x),$ where the input $x$ is the output of the reservoir loop of size N. The FC layers in $Net_a(\cdot)$ are each of size $N\times Q/2$. The bias parameters of the FC layers are set to zero. We add a normalization layer to each FC layer. The applied normalization  must be {\em Layer Normalization}, not batch normalization. Finally, we concatenate the outputs of the 2 layers and apply a ReLU layer, or a Softmax layer, depending on the loss function used for training. We can use either the cross-entropy loss $L_{CE}$ or the MSE (mean square error) $L_{MSE}$.
With Softmax we use $L_{CE}\paren{Net_a(X),Y}$, while with ReLU we use $L_{MSE}\paren{Net_a(X),X}.$ With transfer only, and no further training, we obtained the average accuracy of 98.7 \% using the independent testing set, obtained by merging the two disjoint sets of testing state vectors $X_t = \bigcup{\hat{X^t_i},\ i \in \curls{1,2}}.$ Note that state vectors here are of size N=1000, obtained with the FFT transforms and 10 split loops.

\begin{figure}[h]

\centering
\hspace{-1mm}\includegraphics[width=0.51\textwidth]{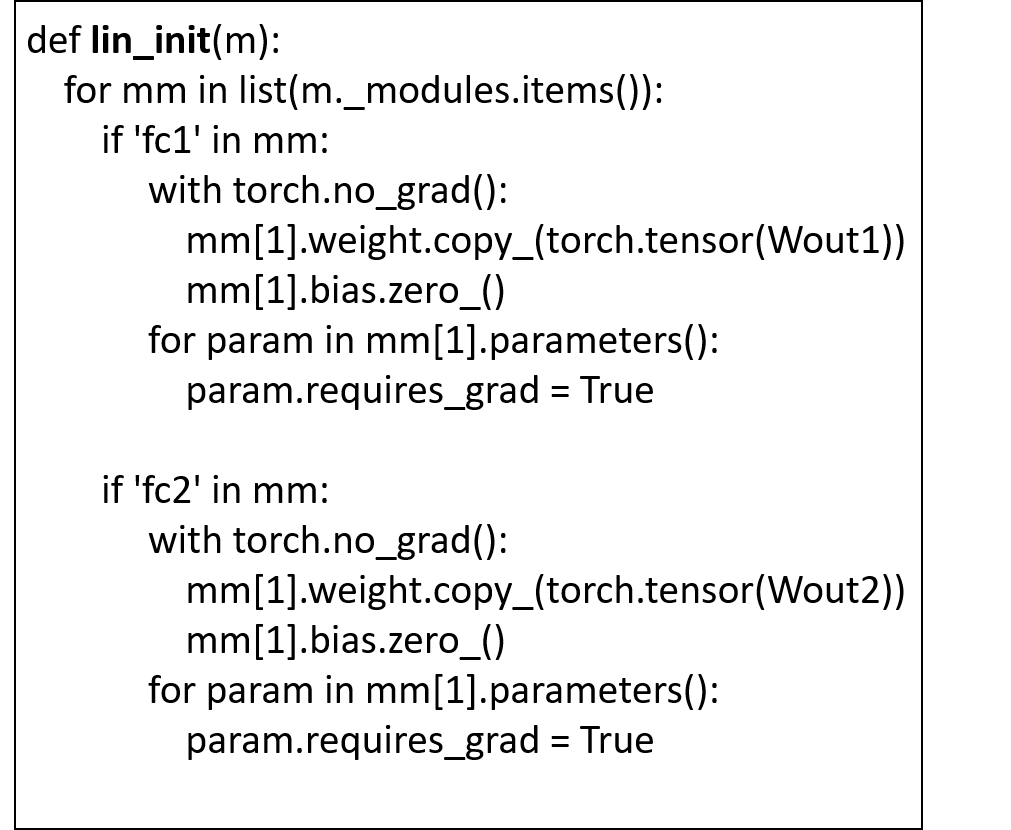}
\vspace{-3mm}
\caption{ Torch-based neural net initialization with $W_{out1/2}$ for the overlapping sets in \eqnref{methodb}, for Q=20.
}\vspace{-3mm}
\label{fig:netbI}
\end{figure}

\begin{figure}[h]

\centering
\hspace{-1mm}\includegraphics[width=0.51\textwidth]{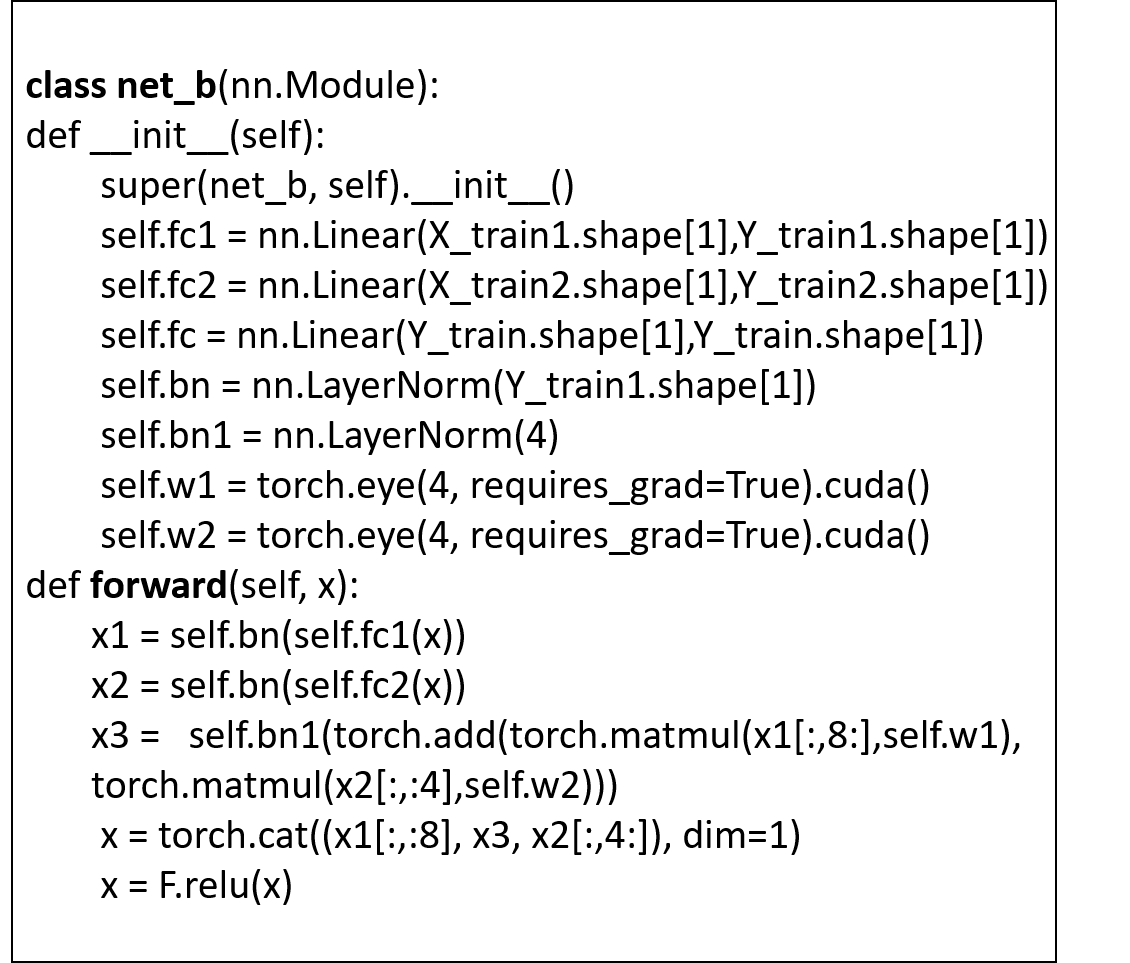}
\vspace{-3mm}
\caption{ Torch-based architecture of the integrated neural net for the overlapping sets in \eqnref{methodb}, for Q=20.
}\vspace{-3mm}
\label{fig:netbA}
\end{figure}

With overlapping  sets of labels, we have 
\begin{align}
\nonumber Y_1=\curls{1,\cdots,3Q/5},\\
Y_2 \in {3Q/5+1,\cdots,Q}.\eqnlabel{methodb}
\end{align}
The respective locally trained weights are obtained as in \eqnref{locals}, except that $\hat{X_i},\  i \in\curls{1,2}$ are the reservoir state vectors obtained from the overlapping sets of training datapoints matching the sets of labels. With $Q=20,$ we combine $W_{out1}$ and $W_{out2}$ by transferring them to the weights of the two fully connected (FC) layers according to the color-coded pattern in Fig.~\ref{fig:distovlp}. With respect to $Net_a(x),$ $Net_b(x)$ has an additional layer (matmul layer in Figs.~\ref{fig:netbI}~and~\ref{fig:netbA}, or the middle FC layer in Fig~\ref{fig:distovlp}): this layer just averages the weights transferred from local models in places that correspond to overlapping devices. This addresses issues associated with unbalanced numbers of training examples in local training sets. We are using equal numbers for now, but it is easy to change the coefficients in the matmul layer to reflect these numbers.
$Net_b(x)$ is using the same loss functions, and same testing sets as $Net_a(x),$ and is achieving about the same average accuracy.
Note that for retraining we use half as many training examples as for the local DLRs in both cases, and we achieve up to 1\% increase.

Finally, note that each DLR can be transferred to its own neural network at no cost apart from the copying of weights, and with no performance decrease.
\section{Cost Calculation for Decentralized Integration of Individual DLR Models}
Let us assume that the total number of identifiable devices is $n.$ Let's denote the number of devices in the $p$th DLR as $n_p,$ and number of devices in the $q$th DLR as $n_q.$ For simplicity, we assume that all devices are mobile, and that there is a significant probability that DLR $p$ will see the devices utilized to train DLR $q$, and vice versa. Hence, the DLRs have an incentive to integrate weight matrices trained by peer DLRs with their own, locally trained ones. We now discuss the communication and computation cost per DLR due to this integration. 
\subsection{Scenario for Fully Inclusive Models} 
Fully inclusive DLR models are needed to support global SEI without some sort of hand-off, which would likely involve tracking individual devices in order to capture when the device might be exiting a local SEI domain and entering the domain of a neighboring DLR. In this fully inclusive case,  the $p$th DLR would need to receive weights for the $n-n_p$ devices covered by others, which involves much more than $n-n_p$ weight vectors, due to forwarding. To make the cost calculation more tractable, let us assume that the global SEI only includes local neighborhoods, i.e., there is no relaying when weights are shared. Let us assume there are $L$ DLRs in the neighborhood. The average cost of sharing the weights (per DLR) is $bN/L\sum_{i=1}^{L}{n_i},$ where $b$ is the number of bytes per weight coefficient. In our case, digital reservoir is implemented with 32-bit floating point math, and the photonic one outputs the same precision through a DAC. The computational complexity beyond local training is negligible as the simple copying of the weights results in the accuracy above 98\% for two merged disjoint subsets of 20 devices. As noted above, we may continue training the resulting neural network to achieve up to 1\% improvement in less than 50 epochs. In many cases that may be considered a negligible improvement and not worth additional cycles, unless it comes as a side effect of transfer-learning in a dynamic environment. Please see Section~\ref{sec:conclusion} for a discussion on the effects of the propagation channel and the receiver variations. 

Our experiments showed that $50\%$ or less of the initial training examples are needed for retraining. Hence, the per-DLR communication cost of retraining the fully inclusive model would be $O\paren{0.5bNB/L}.$ The computational cost of retraining (in number of multiplications) is roughly $C_R = O\paren{\ell_l ENQB},$ where $E$ is the number of epochs, and $\ell_l>1$ is a small constant that depends on the loss function. This cost includes the multiplications in both the forward pass $(0.5ENQB),$ and for computing the Jacobian of the loss  with respect to the weights. Hence, for 50 epochs, N=1000, Q=20, and $B=20*600,$ $C_R = 12e^9\ell_l,$ which is unjustified (compared to Fig.~\ref{fig:fom}) for an average increase in accuracy by 1\% unless we want to perform transfer learning in a dynamic environment. Note that with this additional cost, the total computational cost is still lower than the SoA.
\subsection{Incremental Integration of Uncovered Devices}
We have so far assumed that the training data also includes device labels. This may be justified, e.g., when  legitimate devices want to get covered by an SEI based authentication system. However, in many important cases that is not the case.  Let us see what happens when a rogue device enters the wireless range of a DLR device. The DLR is trained to recognize $Q$ particular devices and it outputs one of the $Q$ labels during the inference phase. Hence, upon capturing the rogue's waveform samples, DLR outputs a random label, which is obviously wrong. How do we identify the SEI outliers - new devices for which the DLR is not trained - to discard the classification results in such cases? This includes mobile devices that are covered by other DLR and not integrated yet, or the new devices that are not part of the DLR system, both friendly and unfriendly. We perform a statistical test to detect outliers, which is based on a statistic derived from the neural nets in Fig.~\ref{fig:distdisj} or Fig.~\ref{fig:distovlp}, or from the local neural net with transferred weights. We calculate the entropy of the Softmax layer outputs, where each of the $Q$ outputs have a value between 0 and 1, and their sum is equal to one, allowing us to treat them as probability distributions. For each input $x$ we calculate the empirical entropy of such a distribution:
\begin{align}
H(x) = \sum_{i=1}^{Q}{p_i(x)\log{p_i(x)}} 
\end{align} 
and we use $H(x)$ in a statistical test to decide if $x$ is a legitimate input or an outlier. The threshold that decides what is an outlier is based on the largest value of $H$ for the training samples.

\begin{figure}[h]
\vspace{-1mm}
\centering
\hspace{-4mm}\includegraphics[width=0.46\textwidth]{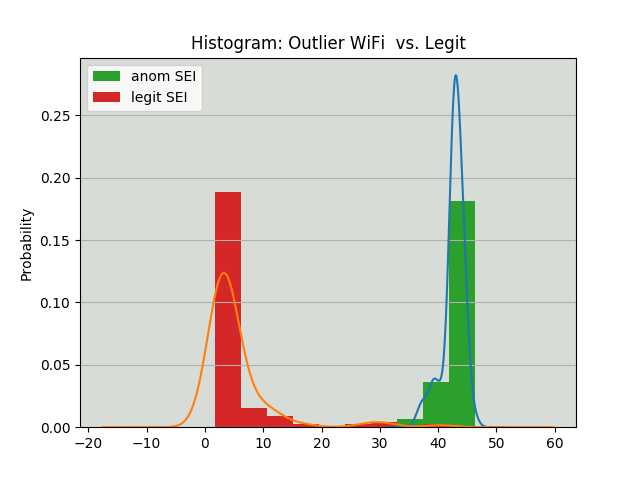}
\vspace{-3mm}
\caption{Histogram of entropy statistics for 100 legitimate and 100 outlier devices: we see that the respective probability distributions have almost disjoint ranges, which is also confirmed by the False positive statistics: only 1\% for both FP95 and FP99.
}\vspace{-3mm}
\label{fig:outlhist}
\end{figure}

The utilization of this statistic results in a very small false positive rate of less than 1\% when true positive is 100\%, equally in the 95 and 99 percentile. We used ROC \cite{ROC} curves to calculate these values. 
Fig.\ref{fig:outlhist} supports these results by showing the histogram  of the entropy statistic for 100 legitimate state vectors and 100 outliers. We obtained legitimate inputs from the set $X_1,Y_1=\curls{1, \cdots, Q/2}$ and the outliers from the set $X_2,Y_2=\curls{Q/2+1, \cdots, Q},$ and we used the neural network transferred from $W_{out1}.$

Hence, we can reliably use the entropy statistic to identify unknown devices and discard the SEI classification in such cases.
\section{Conclusion} \label{sec:conclusion}
There are two major contributions in this work. One is an extension of the prior work that uses our design of   reservoir delay loops (DL) to implement low-power, high accuracy and high-reliability classifiers of signals represented as time series of samples, capable of in-situ training at the Edge. The reported performance figures are based on our H/W implementation of this concept in an FPGA, which we refer to as digital DLR, and on the radio frequency spectrum samples that are collected in a controlled environment, hence no propagation channel analysis is included. That said, we did successfully demonstrate our DLR platform in a live Specific Emitter Identification (SEI) experiment, which included receiver imperfections and some level of white noise. While the demo platform uses both digital and photonic realizations of delay loops, we here only presented the digital one, with several architectures based on trees of DLs. Combining the splitting and multi-layer loops with different degrees of asymmetry between inner and outer DLs, where both inner and outer loop are read out and concatenated or combined, we achieved various levels of  trade-off between the accuracy and complexity of DLR. We demonstrated the effectiveness of the approach on the SEI application, and showed its diversity using Wireless Protocol Recognition (WiPRec). What is evident is that delay loop manages to project the SEI and WiPReq inputs into a space where different input classes are linearly separable. This allows the use of a linear classifier only (such as Ridge Regression or a single layer perceptron) to achieve accuracy much higher than what is possible with the complex non-linear classifiers used in the SoA SEI. Our SEI classifier can be trained orders of magnitude faster on small mobile platforms achieving  remarkable accuracy. 

The second major contribution is distributed training for global SEI-based authentication of mobile devices. This is important for the mobile devices that want to be seamlessly authenticated without the roaming overhead when moving away from the wireless range of their current SEI authentication server. We showed that exchanging weights of locally trained DLRs suffices to maintain accuracy of authentication while expanding the scope. This requires no additional computational cost if the integration is done by transferring the weights into a single layer perceptron in a clean and transparent way. This seamless integration of locally trained sub-models is possible because sub-models are linear classifiers, which is sufficient thanks to using the reservoir. Such a simple neural net is also useful if transfer learning is needed due to dynamic propagation environments. The neural net based outlier detection is effective for denying the SEI-based authentication to devices not previously seen by the global DLR.
Future work includes new applications and additional exploration of combining the loops.  Training on channel-corrupted data is also of interest. As we are currently augmenting the data to quantify the effects of imperfections due to propagation and receiver conditions, an extended version of DLR is appearing to be robust even to fading. The effect of the imperfections by the multiple receivers is also under investigation.

{\bf ACKNOWLEDGMENT: }This research was partially funded by DARPA. The views and conclusions contained in this document are those of the authors and should not be interpreted as representing the official policies, either expressed or implied, of the U.S. Government. 
DISTRIBUTION STATEMENT A. Approved for public release: distribution unlimited
\bibliographystyle{IEEEtran}%
\bibliography{DLRbasic}

\begin{thebibliography}{10}
\providecommand{\url}[1]{#1}
\csname url@samestyle\endcsname
\providecommand{\newblock}{\relax}
\providecommand{\bibinfo}[2]{#2}
\providecommand{\BIBentrySTDinterwordspacing}{\spaceskip=0pt\relax}
\providecommand{\BIBentryALTinterwordstretchfactor}{4}
\providecommand{\BIBentryALTinterwordspacing}{\spaceskip=\fontdimen2\font plus
\BIBentryALTinterwordstretchfactor\fontdimen3\font minus
  \fontdimen4\font\relax}
\providecommand{\BIBforeignlanguage}[2]{{%
\expandafter\ifx\csname l@#1\endcsname\relax
\typeout{** WARNING: IEEEtran.bst: No hyphenation pattern has been}%
\typeout{** loaded for the language `#1'. Using the pattern for}%
\typeout{** the default language instead.}%
\else
\language=\csname l@#1\endcsname
\fi
#2}}
\providecommand{\BIBdecl}{\relax}
\BIBdecl

\bibitem{RFSoC}
\BIBentryALTinterwordspacing
Xilinx, ``{Zynq UltraScale+ RFSoC ZCU111 Evaluation Kit},'' accessed on
  10/7/2020. [Online]. Available:
  \url{https://www.xilinx.com/products/boards-and-kits/zcu111.html}
\BIBentrySTDinterwordspacing

\bibitem{AdvancesinphotonicRC}
G.~V. der Sande, D.~Brunner, and M.~C. Soriano, ``Advances in photonic
  reservoir computing,'' \emph{Nanophotonics}, vol.~6, no.~3, pp. 561 -- 576,
  01 May. 2017.

\bibitem{Lukosevicius2009ReservoirCA}
M.~Lukosevicius and H.~Jaeger, ``Reservoir computing approaches to recurrent
  neural network training,'' \emph{Comput. Sci. Rev.}, vol.~3, pp. 127--149,
  2009.

\bibitem{GomacTech}
{S. Kokalj-Filipovic and P. Toliver and W. Johnson et al.}, ``Deep delay loop
  reservoir computing for specific emitter identification,'' \emph{arXiv
  preprint: 2010.06649}, 2020, appeared in GOMACTech 2020.

\bibitem{ICCpaper}
{S. Kokalj-Filipovic and P. Toliver and W. Johnson and R. Miller}, ``{Reservoir
  Based Edge Training on RF Data To Deliver Intelligent and Efficient IoT
  Spectrum Sensors},'' in \emph{Submitted to IEEE SmartComp}, 2021.

\bibitem{80211n}
IEEE, ``{IEEE P802.11n™/D3.00 - Draft Amend. to STANDARD for ITT and inform.
  exchange between Systems - Local and Metropolitan networks-Specific reqs-Part
  11: Wireless LAN MAC and PHY. Amendm. 4: Enhancements for Higher
  Throughput}.''

\bibitem{BT}
\BIBentryALTinterwordspacing
B.~S.~I. Group, ``{Bluetooth Specification Version 5.0},'' accessed on
  10/25/2018. [Online]. Available:
  \url{https://www.bluetooth.com/specifications/bluetooth-core-specification}
\BIBentrySTDinterwordspacing

\bibitem{ZBee}
N.~Salman, I.~Rasool, and A.~H. Kemp, ``{Overview of the IEEE 802.15.4
  standards family for Low Rate Wireless Personal Area Networks},'' in
  \emph{7th Int. Symp. on Wireless Communication Systems}, 2010.

\bibitem{NRF}
\BIBentryALTinterwordspacing
N.~Semiconductor, ``{nRF24L01 Single Chip 2.4GHz Transceiver},'' 2018, accessed
  on 10/19/2020. [Online]. Available:
  \url{https://www.sparkfun.com/datasheets/Components/nRF24L01_prelim_prod_spec_1_2.pdf}
\BIBentrySTDinterwordspacing

\bibitem{SDAE}
S.~{Kokalj-Filipovic}, R.~{Miller}, and J.~{Morman}, ``Autoencoders for
  training compact deep learning rf classifiers for wireless protocols,'' in
  \emph{IEEE Intl. Workshop on Signal Processing Advances in Wireless Comms
  (SPAWC)}, 2019, pp. 1--5.

\bibitem{CabricSEIAuthor}
S.~Hanna, S.~Karunaratne, and D.~Cabric, ``Open set wireless transmitter
  authorization: Deep learning approaches and dataset considerations,'' 2020.

\bibitem{SEIIoTJnl}
J.~M. {McGinthy}, L.~J. {Wong}, and A.~J. {Michaels}, ``Groundwork for neural
  network-based specific emitter identification authentication for iot,''
  \emph{IEEE Internet of Things Journal}, vol.~6, no.~4, 2019.

\bibitem{PUF}
B.~{Chatterjee}, D.~{Das}, S.~{Maity}, and S.~{Sen}, ``{RF-PUF: Enhancing IoT
  Security Through Authentication of Wireless Nodes Using In-Situ Machine
  Learning},'' \emph{IEEE Internet of Things Journal}, vol.~6, no.~1, pp.
  388--398, 2019.

\bibitem{deepunfolding}
A.~{Balatsoukas-Stimming} and C.~{Studer}, ``Deep unfolding for communications
  systems: A survey and some new directions,'' in \emph{2019 IEEE Intnl
  Workshop on Signal Proc. Systems (SiPS)}, 2019.

\bibitem{modelbased}
H.~{He}, S.~{Jin}, C.~{Wen}, F.~{Gao}, G.~Y. {Li}, and Z.~{Xu}, ``Model-driven
  deep learning for physical layer communications,'' \emph{IEEE Wireless
  Communications}, vol.~26, no.~5, pp. 77--83, 2019.

\bibitem{policybased}
{R. Miller et al.}, ``{POLICY BASED SYNTHESIS: Data Generation and Augmentation
  Methods for RF Machine Learning},'' in \emph{IEEE GlobalSIP}, Nov 2019.

\bibitem{PhotRCTut}
D.~Brunner, B.~Penkovsky, and B.~A.~M. et~al., ``{Tutorial: Photonic neural
  networks in delay systems},'' \emph{Journal of Applied Physics}, 2018.

\bibitem{RR}
A.~E. Hoerl and R.~W. Kennard, ``Ridge regression: Biased estimation for
  nonorthogonal problems,'' \emph{Technometrics}, vol.~12, no.~1, pp. 55--67,
  1970.

\bibitem{SKay}
S.~Kay, ``{A Fast and Accurate Single Frequency Estimator},'' \emph{{IEEE
  Trans. on Acoutics Speech and Signal Proc.}}, vol.~37, no.~12, 1989.

\bibitem{pseudoinverse}
\BIBentryALTinterwordspacing
V.~Pan, ``Complexity of computations with matrices and polynomials,''
  \emph{SIAM Review}, vol.~34, no.~2, pp. 225--262, 1992. [Online]. Available:
  \url{http://www.jstor.org/stable/2132854}
\BIBentrySTDinterwordspacing

\bibitem{ROC}
{J.P. Egan}, \emph{{Signal Detection Theory and ROC Analysis}}.\hskip 1em plus
  0.5em minus 0.4em\relax Academic Press, New York, 1975.

\end{thebibliography}
\end{document}